%% file: tacl.tex
\documentclass[11pt,letterpaper]{article}
\usepackage[letterpaper]{geometry}
\usepackage{acl2012}

\usepackage{times}
\usepackage{latexsym}
\usepackage{amsmath}
\usepackage{multirow}
\usepackage{url}
\makeatletter
\newcommand{\@BIBLABEL}{\@emptybiblabel}
\newcommand{\@emptybiblabel}[1]{}
\makeatother

\usepackage{siunitx}

\usepackage{times}
\usepackage{helvet}
\usepackage{courier}
\usepackage{graphicx}
\usepackage{color}
\usepackage{textcomp}
\usepackage{booktabs}
\usepackage{ccicons}
\usepackage{pifont}

\usepackage{verbatim}
\usepackage{multirow}
\usepackage{threeparttable}
\usepackage{booktabs}
\usepackage{graphicx}
\usepackage{amsmath}
\usepackage{mdwlist}
\usepackage{grffile}
\usepackage{graphicx}
\usepackage[font=small]{caption}
\usepackage[font=small]{subcaption}
\usepackage{eurosym}
\usepackage{mdwlist}

\usepackage{multirow}
\usepackage{pbox}
\usepackage{enumerate}
\usepackage{url}
\usepackage{rotating}
\usepackage{color, colortbl}
\usepackage{dcolumn}

\usepackage{hyperref}

\newcommand{\specialcell}[2][c]{%
  \begin{tabular}[#1]{@{}c@{}}#2\end{tabular}}

\title{Citation Classification for Behavioral Analysis of a Scientific Field}

 \author{David Jurgens \\ Stanford University \\ {\tt jurgens@stanford.edu}
         \And Srijan Kumar  \\  University of Maryland, College Park \\ {\tt srijan@cs.umd.edu} 
         \AND Raine Hoover \\ Stanford University \\ {\tt raine@stanford.edu}
         \And Dan McFarland \\ Stanford University \\ {\tt dmcfarla@stanford.edu}
         \And Dan Jurafsky \\ Stanford University \\ {\tt jurafsky@stanford.edu}
}

\date{}

\begin{document}

\maketitle
\begin{abstract}
Citations are an important indicator of the state of a scientific field,
reflecting how authors frame their work, and influencing uptake by future scholars.
However, our understanding of citation behavior has been limited to small-scale manual citation analysis.
We perform the largest behavioral study of citations to date, analyzing how citations are both framed and
taken up by scholars in one entire field: natural language processing.
We introduce a new dataset of nearly 2,000 citations annotated for function
and centrality, and use it to develop a state-of-the-art classifier and label 
the entire ACL Reference Corpus.  We then study how citations are framed by authors
and use both papers and online traces to track how citations are followed by readers. We demonstrate
that authors are sensitive to discourse structure and publication venue when citing,
that online readers follow temporal links to previous and future work rather than methodological links,
and that  how a paper cites related work is predictive of its citation count.
Finally, we use changes in citation roles to show that the field of NLP is undergoing a significant increase in consensus.
\end{abstract}

\input{1-intro}

\input{2-corpus}

\input{3-classifier}

\input{4-venue}

\input{5-browsing}

\input{6-predict-impact}
\input{7-rapid}

\input{9-conclusion}

\bibliography{references}
\bibliographystyle{acl2012}

\end{document}

%% file: 1-intro.tex
\section{Introduction}

Citations play a key role in  scientific development.   The citations from a paper reinforce its arguments and connect it to an intellectual lineage \cite{latour1987science}, while the citations to a paper enable communities  to evaluate its intellectual
contribution and quality \cite{cole1971measuring,Lindsey1980production,hirsch2005index}.
At the same time, authors employ citations in multiple ways (Figure 1) so as to build a strong and multi-faceted argument \cite{latour1987science}. 
While research on citation function is longstanding \cite{swales1986citation,white2004citation,ding2014content},
there is still no field-scale dataset for citation function, and so
we still don't understand the way that a field's authors frame their citations 
and how this framing influences uptake by readers and future citers.  

We perform the first field-scale study of citation usage by developing accurate methods for automatically classifying citation purpose,
and  applying these methods to an entire field's literature.

\begin{figure}
\centering
    \includegraphics[width=0.41\textwidth]{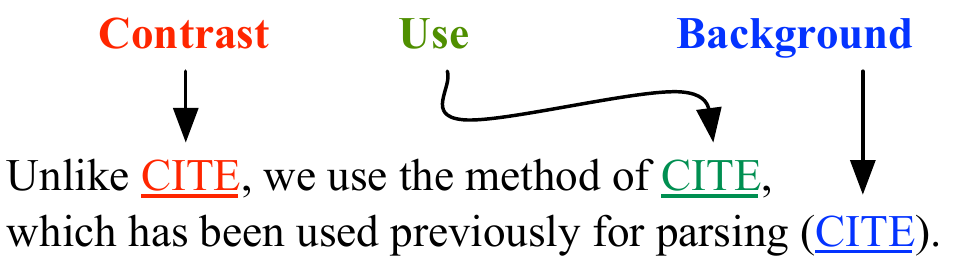}
    \caption{Examples of citation functionality.}
    \label{fig:example}
    \vspace{-2mm}
\end{figure}

We unify core aspects of prior citation annotation schemes \cite{white2004citation,ding2014content,hernandez2016survey}
into a highly-operationalizable coarse-grained classification that captures
the {\em function} the citation plays towards furthering an argument 
as well as whether the citation is {\em essential} for understanding a contribution or serves rather to {\em position} the paper within the scientific context.
Using this scheme, we annotate a corpus of citations and use it to
train a high-accuracy  method for automatically labeling a corpus.
We use this method  to label the field of NLP, over 134,127 citations in over 20,000 papers from nearly forty years of work.

We then investigate a number of key questions concerning how authors frame their citations
(how citations reflect the discourse structure of a paper, how they are influenced by venue)
how readers take them up (which citations do online readers follow, how does citation framing affect
future citations to a paper), and how these processes reflect the maturation and growth of the field of NLP.
As further contributions, we publicly release our dataset and code.

%% file: 2-corpus.tex
\section{A Corpus for Citation Function}
\label{sec:corpus}

Citations play a key role in supporting authors' contributions throughout a scientific paper.\footnote{For notational clarity, we use the term \emph{reference} for the work that is cited and \emph{citation} for the mention of it in the text.}   Multiple schemes have been proposed on how to classify these different roles, ranging from a handful of classes \cite{Nanba1999towards,Pham2003new} to twenty or more \cite{Garfield1979citation,Garzone2000towards}.  
While suitable for expert manual analysis, such schemes don't fit our goal of field-scale automatic classification.
Because they tend to be unidimensional, they can't label both centrality and  function \cite{swales1986citation}, 
they often include fine-grained distinctions too rare to reliably identify and
subjective classifications that require detailed field or author knowledge \cite{ziman1968public,swales1990genre,harwood2009interview}.
Motivated by the desire to examine large-scale trends in scholarly behavior,
we address these issues by unifying the common aspects of multiple approaches in a two-dimensional model.

\begin{table*}[t]
  \centering
  \footnotesize
    \begin{tabular}{@{}p{24mm}@{\hspace*{4pt}} p{1.5in} p{3.8in}}
        \textbf{Class} & \textbf{Description} & \textbf{Example} \\
        \hline
    
      \textsc{Background} & $P$ provides relevant information for this domain. & {This is often referred to as incorporating deterministic closure (D\"{o}rre, 1993).} \\
      \textsc{Motivation} & $P$ illustrates need for data, goals, methods, etc. & {As shown in Meurers (1994) this is a well-motivated convention  [...]}\\
      \textsc{Uses} & Uses data, methods, etc., from $P$ . & The head words can be automatically extracted  [...]
      in the manner described by Magerman (1994).  \\

     \specialcell{\textsc{Extension}} & Extends $P$'s data, methods, etc. & [...] we improve a two-dimensional multimodal version of LDA (Andrews et al, 2009) [...] \\
      \textsc{Continuation} & Expands $P$  by same authors & This section elaborates on preliminary results reported in Demner-Fushman and Lin (2005), [...]\\
      \textsc{Comparison} ~~~~\textsc{or Contrast} & Expresses similarity/differences to $P$. & Other approaches use less deep linguistic resources (e.g., POS-tags Stymne (2008)) [...]\\
      \textsc{Future} & $P$ is a potential avenue for future work. & 
      [...] but we plan to do so in the near future using the algorithm of Littlestone and Warmuth (1992). \\
    \end{tabular}

    \caption{Our set of seven functions a citation may serve
    with respect to a cited  paper $P$.}
    \label{tab:classifications}
\end{table*}

\subsection{Classification Scheme}

Our classification builds on two themes for citation role:
(1) citation centrality, which reflects whether the citation is used to position the contribution within a broader context or is essential for understanding the contribution itself \cite{moravcsik1975some,chubin1975content,swales1986citation,latour1987science,valenzuela2015identifying}
and (2) citation function, which reflects the particular purpose a citation is serving in the discourse, e.g., providing background or serving as contrast \cite{Oppenheim1978highly,Spiegel1977bibliometric,teufel2006annotation,Garfield1979citation,Garzone2000towards}.
These themes capture complementary information and, as we will see, reveal meaningful patterns in author behavior.\footnote{A third potential theme is  citation sentiment \cite{athar2014sentiment,kumar2016structure}, but
we omit this because researchers have shown that  negative sentiment is rare in practice \cite{chubin1975content,vinkler1998comparative,case2000can} and 
quite subjective to classify due to textual mixtures of praise and criticism \cite{peritz1983classification,swales1986citation,brooks1986evidence,teufel2000argumentative}.}
%

\noindent \textbf{Centrality}~~
In citing, an author attempts to persuade the reader of a paper's merit \cite{gilbert1977referencing} by
incorporating other work to further their own approach,
or by  contextualizing their work within the broader literature. 
Both kinds of citations are necessary; for example as \newcite{latour1987science} and others have argued, positioning
citations secure the inductive gaps between an author's 
arguments,  and allow the author to identify with a lineage of work as motivation for their own position.
The first theme in our classification,
\emph{citation centrality}, thus
specifies whether a citation is 
(a) \textsc{Essential} to understanding the contributions of the citing paper or
(b) \textsc{Positioning} by situating the citing paper within a broader context.

\noindent \textbf{Function}~~
Citation 
{\em function} reflects the specific purpose a citation plays with respect to the current paper's contributions.  We unify the functional roles common in several classifications, e.g., \cite{Spiegel1977bibliometric,Garfield1979citation,peritz1983classification,teufel2006annotation,harwood2009interview,dong2011ensemble}, into the seven classes shown in Table~\ref{tab:classifications}.

\subsection{Annotation Process and Dataset}

Annotation guidelines were created using a pilot study of 10 papers sampled from the ACL Anthology Reference Corpus (ARC) \cite{bird2008acl}.  Annotators  completed two rounds of pre-annotation to discuss their process and design  guidelines.
All citations were then doubly-annotated by two trained annotators using
the Brat tool \cite{stenetorp2012brat} and then  fully adjudicated to ensure quality.  

The citation scheme was applied to a random sample of 52 papers drawn from the ARC.  Each
paper was processed using ParsCit \cite{councill2008parscit} to extract 
citations and their references.  
As expected from  prior studies \cite{teufel2006annotation,dong2011ensemble}, some citation functions were infrequent.
We therefore attempted to oversample the infrequent classes 
\textsc{Future}, \textsc{Continuation}, and \textsc{Motivation},
by using keywords biased toward extracting citing sentences of 
a particular class (such as the word ``future''  for the \textsc{Future} class).
The resulting citing sentences  were then annotated and could
potentially be assigned to any class.
In total, 1436 contexts were annotated for the fully-labeled 52 papers  (mean 27.6 citations/paper) and 533 supplemental contexts were added by sampling, bringing the total number of instances to 1969.

Table \ref{tab:ann-data} shows the final dataset.  Consistent with prior work, the majority of citations are \textsc{Background} \cite{moravcsik1975some,Spiegel1977bibliometric,teufel2006automatic}.  
While some citation functions are strongly associated with one centrality type, e.g., \textsc{Background} as \textsc{Positioning}; citation function does not wholly predict centrality, highlighting the need for the two complementary classifications.

\begin{table}
    \small
     \begin{tabular}{l @{\hspace*{4pt}}c@{\hspace*{4pt}}c@{\hspace*{4pt}}c}
        \textbf{Citation Function} & \textbf{Count} & \textbf{\% Positioning} & \textbf{\% Essential} \\
        \hline
        \textsc{Background} & 1021 & 1.0~~ & 0.0~~ \\
        \textsc{Uses} & ~~365 & 0.02 & 0.98 \\
        \textsc{Compares } & ~~344 & 0.77 & 0.23 \\
        \textsc{\ \ or Contrasts} & \\
        \textsc{Motivation} & ~~~~98 & 0.93 & 0.07 \\
        \textsc{Future} & ~~~~68 & 1.0~~ & 0.0~~ \\
        \textsc{Continuation} & ~~~~51 & 0.10 & 0.90 \\
        \textsc{Extends} & ~~~~22 & 0.05 & 0.95 \\
    \end{tabular}
    \caption{Citation class distribution in the dataset}
    \label{tab:ann-data}
    \vspace{-3mm}
\end{table}

%% file: 3-classifier.tex
\section{Automatically Classifying Citations}
\label{sec:classifier}

The structure of a scientific article provides multiple cues for a citation's purpose.  Our work draws on multiple approaches \cite{hernandez2016survey} to develop a classifier based on  (1) \emph{structural} features describing where the citation is located, (2) \emph{lexical}  and \emph{grammatical} features for how the citation is described,  (3) \emph{field} features that take into account venue or other external information, and (4) \emph{usage} features on how the  reference is cited throughout the paper.  
Table \ref{tab:features} shows our features, including
those drawn from state of the art systems \cite{teufel2000argumentative,teufel2006automatic,dong2011ensemble,wan2014all,valenzuela2015identifying}
as well as ten novel feature classes.

\begin{table}[t]
    \small
    {
    \begin{tabular}{l}
        \hline
       {\bf Structural}\\ 
        \hline
    section \# and remaining \# of sections \\
    relative positions in paper, section, subsection\\
    ~~~~~~sentence, \& clause \\
    \# of other citations in subsection, sentence, \& clause\\
    canonicalized section title \\
    \hline
       {\bf Lexical, Morphological, and Grammatical}\\ 
        \hline
    function patterns of \newcite{teufel2000argumentative} \\
    $\dagger$ bootstrapped function patterns \\
    $\dagger$ custom function patterns \\
    $\dagger$ citation prototypicality \\
    $\dagger$ citation context topics \\
    $\dagger$ paper topics \\
    topical similarity with cited paper \\
    the presence of each of 23 connective phrases \\ 

 
    $\dagger$ whether used in nominative or parenthetical form, \\ 
    $\dagger$ whether preceded by a Pascal-cased word \\
    $\dagger$ whether preceded by an all-capital case word  \\
    verb tense \\ 
    lengths of the containing sentence and clause \\
        whether used inside of a parenthetical statement  \\

   \hline
   {\bf Field}\\
        \hline
   $\dagger$ citing paper's venue: journal/conference/workshop \\ 

   $\dagger$ reference's venue: journal/conference/workshop \\ 
   reference's citation count and PageRank \\
    ~~~~~~(at time of the citation) \\
   \# of years difference in publication dates \\
   whether the cited paper is a self-citation \\
   \hline
   {\bf Usage}\\
        \hline
   \# of indirect citations  \\ 
   \# of direct citations \\
   \# of indirect citations per section type\\
   \# of direct citations per section type \\
   fraction of bibliography used by this reference \\[-6pt]
   \end{tabular}
   }
   \caption{Features for classifying citations. Novel features are marked with a $\dagger$.}
   \label{tab:features}
\end{table}

\subsection{Features}

Following, we describe in detail the three main categories of novel features.

\noindent \textbf{Pattern-based Features}~~
Patterns provide a powerful mechanisms for capturing regularity in citation usage  \cite{dong2011ensemble}. Our patterns are a sequence of cue phrases, parts of speech, or lexical categories, like positive-sentiment words or
specific categories that allow generalizations  across phrases like ``we extend'' and ``we build upon.'' We began with the largest
publicly-available list of citation  patterns 
\cite{teufel2000argumentative}, extending it
with 132 new patterns and 13 new lexical categories based on a manual analysis of the corpus.
%
%

We then used bootstrapping to automatically identify new patterns.
Each annotated context was converted into fixed-length patterns using (a) our 42 lexical categories, (b) part of speech wild cards, or (c) the tokens directly.
To avoid semantic drift \cite{RiloffJones99},
a bootstrapped pattern was only included as a feature if the majority of its occurrences were with a single citation function.\footnote{For computational efficiency, patterns were restricted to having between 3 and 8 tokens and at most two part of speech wild cards.  Due to its high frequency, patterns for \textsc{Background} were required to occur in at least 100 contexts.}  Table \ref{tab:bootstrapped} shows examples of these bootstrapped patterns. 

\begin{table}[t]
    \centering
    \footnotesize
        {
        \begin{tabular}{l@{\hspace*{3pt}}l}
    \textbf{Function} & \textbf{Pattern} \\ 
    \hline
     \textsc{Comp.~or~Con.} & @{\sc similar\_adj} to @{\sc referential} @{\sc use}   \\ 
     \textsc{Comp.~or~Con.} &  the @{\sc Research\_Noun} of \#N           \\ 

    \textsc{Extends} &       @{\sc Change\_Noun} of \#N 's    \\ 
    \textsc{Extends} &        @{\sc Change\_Noun} of \emph{citation} 's     \\ 

    \textsc{Motivation} & @{\sc Inspiration} by \#N           \\ 

    \textsc{Continuation} & @{\sc Poss\_Pronoun} @{\sc Before\_Adj} @{\sc Work\_Noun}           \\ 
    \textsc{Continuation} & @{\sc Before\_Adj} @{\sc Work\_Noun} \#N         \\ 

    \textsc{Uses} &  @{\sc 1st\_Person\_Pronoun\_(Nom)} @{\sc Use} the \#N      \\ 
    \textsc{Uses} & the \#N corpus       \\ 
    \textsc{Uses} & \#D \#N \#N \emph{citation}        \\ 

        \end{tabular}
    }
    \caption{Examples of bootstrapped patterns learned and their associated class where @ denotes a lexical class and \# denotes a part of speech wild card.}
    \label{tab:bootstrapped}
\end{table}

Previous patterns primarily use cues from the same sentence as the citation
\cite{teufel2000argumentative}.
However, authors often use multiple
sentences to indicate a citation's purpose \cite{abu2012reference,ritchie2008comparing,he2011citation,kataria2011context}.
For example, authors may first introduce a work positively, only to contrast with it in later sentences \cite{peritz1983classification,brooks1986evidence,mercer2004frequency}.
Indeed the average text pertaining to a citation spans 1.6 sentences
in the ARC \cite{small2011interpreting}.
%

We therefore induce bootstrapped patterns specific to the citation sentence as well as the preceding and following sentences.
%
Ultimately, 805 new bootstrapped patterns were added for the citing sentence, 669 for the preceding context, and 1159 for the following, over four times the number of manually curated patterns.
%

\noindent \textbf{Topic-based Features}~~
A context's thematic framing can point to the purpose of a citation even in the absence of explicit cues.  For example, a context describing system performances and results is likely to be a \textsc{Compare or Contrast}, whereas one describing methodology is more likely to be \textsc{Uses}.  
We quantify this thematic framing by using features based on topic models, 
computed over
the sentence containing the citation and also over an extended \mbox{-1/+3} context around the citing sentence.
For each type of context, a topic model is trained with 100 topics over 321,129 respective contexts from the ARC.  Table \ref{tab:topics} shows example topics.

\begin{table}[t]
    \small 
    \centering
    \resizebox{0.51\textwidth}{!}{
    \begin{tabular}{l}
    \hline
     \hspace{-3mm} 1) algorithm parameter model training method clustering   \\
   \hspace{-3mm} 2) measure score metric information similarity distance   \\
    \hspace{-3mm} 3) \% result accuracy report achieve performance system   \\
   \hspace{-3mm} 4) training weight feature och model set algorithm error  \\
    \hspace{-3mm} 5) work related previous paper problem approach present   \\
    \hline
    \end{tabular}
    }
    \caption{The most probable words from five example topics learned from citation contexts.}
    \label{tab:topics}
\end{table}

\noindent  \textbf{Prototypical Argument Features}~~
We also explored richer grammatical features, drawing on
selectional preferences reflecting expectations for predicate arguments.
We construct a prototype for each citation function by identifying the frequent arguments seen in different syntactic positions.
For example, \textsc{Continuation} citations occur frequently as objects of verbs such as ``follow'' and ``use,'' whereas \textsc{Uses} citations have techniques or artifact words as dependents; Table~\ref{tab:dep-examples} shows more examples.
Each class's selectional preferences are represented using a vector for the argument at each relation type, constructed by summing the vectors of all words appearing in it.  
Each function is represented as a separate feature whose value is the  average similarity of an instance's arguments with the class's preferences for all observed syntactic relationships (i.e., how similar are the syntactically-related words to the function's preferences).

\begin{table}[t]
    \small 
    \resizebox{0.52\textwidth}{!}{  
    \begin{tabular}{ @{}l@{\hspace*{3pt}}l@{\hspace*{3pt}}l }
        \textbf{Function} & \textbf{Path} & \textbf{Arguments} \\
            \hline
\textsc{Motivation}      &     nmod$^{-1}$ &  inspire, work, show\\
\textsc{Motivation}      &     nmod$^{-1}$, nmod$^{-1}$     &   exemplify, direction, inspire\\

\textsc{Continuation}   &      amod  &     previous, prior, unsupervised\\
\textsc{Continuation}   &     dobj$^{-1}$  & follow, use, describe\\

\textsc{Uses}    &     nmod$^{-1}$ &  use, describe, propose\\
\textsc{Uses}    &     dep$^{-1}$  &  system, algorithm, mechanism\\

\textsc{Comp. or Cont.}       &      nmod$^{-1}$, nmod$^{-1}$  &     similar, related, use\\
\textsc{Comp. or Cont.}       &      dep$^{-1}$   & system, method, approach\\

\textsc{Extends} &      nmod$^{-1}$, nmod$^{-1}$  &      base, version, extension \\
\textsc{Extends} &       dobj$^{-1}$  & follow, extend, unfold \\
    \end{tabular}
    }
    \caption{Examples of citation function selectional preferences with the most-frequent words in paths.}
    \label{tab:dep-examples}

\end{table}

\subsection{Experimental Setup}

\noindent \textbf{Models}~~
All models were trained using a Random Forest classifier, which is robust to overfitting even with large numbers of features 
\cite{fernandez2014we}.
After limited grid search, we set the number of random trees to 500, initialized trees using a balanced subsample of the data, and required each leaf to match 7 instances.   The classifier is implemented using SciKit \cite{pedregosa2011scikit} 
and  syntactic processing was done using  CoreNLP \cite{manning2014stanford}.  Selectional  preferences used pretrained 300-dimensional vectors from the 840B token Common Crawl \cite{pennington2014glove}.

\noindent \textbf{Data}~~
Annotated data is crucial for developing high accuracy for rare
citation classes.  Therefore, we integrate portions of the dataset
of \newcite{teufel2010structure}, which has fine-grained citation
function labeled for ACL-related documents using the annotation
scheme of \newcite{teufel2006automatic}.  We map their 12 function
classes into six of ours.   When combining the two datasets, we
omit the data labeled with their \textsc{Background}-equivalent
class to reduce the effects of a large majority class and because
instances of the \textsc{Future} class are merged into \textsc{Background}
according to their scheme.  The resulting citation function dataset
contains 3083 instances.  As there is no precise mapping from
function to centrality (cf.~Table \ref{tab:ann-data}), centrality
classifiers use only our annotated data.

\noindent \textbf{Evaluation}~~
Evaluation is performed using cross-validation where each fold leaves out all citations of a single paper.  Stratifying by paper instead of instance is critical: since multiple citations may appear in the same sentence, instance-based stratification would leak information between training and test.  
We report micro- and macro-averaged F1 scores across the seven function classes and for centrality, report precision and recall on the binary classification where \textsc{Essential} is the positive class.

\noindent \textbf{Comparison Systems}~~
The proposed citation classifiers are compared against three systems.  For state of the art, we compare against \newcite{teufel2000argumentative} which is the most similar model to ours that is experimentally reproducible; the original implementation used a custom syntactic tool (for e.g., verb tense), which we replaced with  CoreNLP. 
Two baselines are used for comparison: a Random baseline that selects labels at chance and a Single-class
baseline that labels all instances with the most frequent citation function \textsc{Background} or in the binary centrality classification, with the positive class \textsc{Essential}.

%

\begin{table}[t]
    {
    \footnotesize
  \begin{tabular}{@{}l@{\hspace*{2pt}}c@{\hspace*{2pt}}c @{\hspace*{6pt}}l@{\hspace*{2pt}} c@{\hspace*{2pt}}c}
    & \multicolumn{2}{c}{Function} &&  \multicolumn{2}{c}{Centrality} \\
    &  Micro F1 &  Macro F1 && Precision & Recall \\
    \hline
    This work                                 & \textbf{0.592} & \textbf{0.498}    && \textbf{0.684} & 0.562 \\ 
\ \ \emph{no bootstrapped pats.}  & 0.588 & 0.498    && 0.674 & 0.560 \\ 
\ \ \emph{no custom patterns}     & 0.582 & 0.469    && 0.676 & 0.548 \\ 
\ \ \emph{no topic features}      & 0.579 & 0.468    && 0.633 & 0.633 \\ 
\ \ \emph{no selectional prefs.}  & 0.572 & 0.469    && 0.675 & 0.568 \\ 
\ \ \emph{no novel features}      & 0.569 & 0.464    && 0.625 & 0.644 \\ 
    \newcite{teufel2000argumentative}         & 0.370 & 0.248    && 0.470 & 0.422 \\
    One-Class                                 & 0.331 & 0.071    && 0.259 & \textbf{1.000} \\
    Random                                    & 0.134 & 0.107    && 0.253 & 0.491 \\

%
%
    
  \end{tabular}
  }
  \caption{Classifier performances.}
  \label{tab:perf}
\end{table}

\subsection{Results and Discussion}

Our methods substantially outperformed
the closest state of the art and both baselines for both classification tasks, as shown in Table \ref{tab:perf}. 
All  improvements over comparison systems are statistically significant (McNemar's, p$\le$0.01).

An ablation test suggests that each of our novel features contributed to the final performance.  Notably, we observe that selectional preference and topic features had the largest impact on performance.  While both we and multiple prior works have focused on patterns to recognize function, our results suggest that machine learned features (topics or word vectors) are superior.
Indeed, examining the feature weighting in the random forest shows that features for structure (e.g., section number), topic, and selectional preference comprised most of the 100 highest-weighted features (76\%).

The use of conjunctive features was critical for performance, with all other classifiers  we tried
(SVM, Naive Bayes, $k$-nearest neighbor, and Decision Trees)
providing significantly worse results.\footnote{Indeed, replacing the $k$-nearest neighbors classifier used in \newcite{teufel2000argumentative} with a random forest improves citation function classification by 0.123 (Macro F1) and 0.175 (Micro F1).}

The resulting classifier performance is sufficient to apply it to
the entire ARC dataset for the analyses  in the next four sections.
Nonetheless, errors remain.  Our error analysis revealed that a main challenge is incorporating information external to the citing sentence.
Consider the following example:
\begin{list}{
   \setlength{\rightmargin}{0pt}}
     \item
\vspace*{-10pt}
     \footnotesize
BilderNetle is our new data set of German noun-to-ImageNet synset mappings. ImageNet is a large-scale and widely used image database, built on top of WordNet, which maps words into groups of images, called synsets (Deng et al., 2009).
\end{list}

Here the citing sentence appears much like a \textsc{Background} citation when read in isolation; however, the preceding sentence reveals that the citing work's data is based on the citation, making its function \textsc{Uses} though no explicit cues suggest this in the citing sentence.
Automated methods need to do a better job of inferring  (a) which context
inside and outside the citing sentence relates to the citation and
(b) how the cited paper's content relates to this context; together
these require  richer  textual understanding than is present in
most current methods.

\begin{figure}[t]
    \hspace*{-14pt}
    \includegraphics[width=0.55\textwidth]{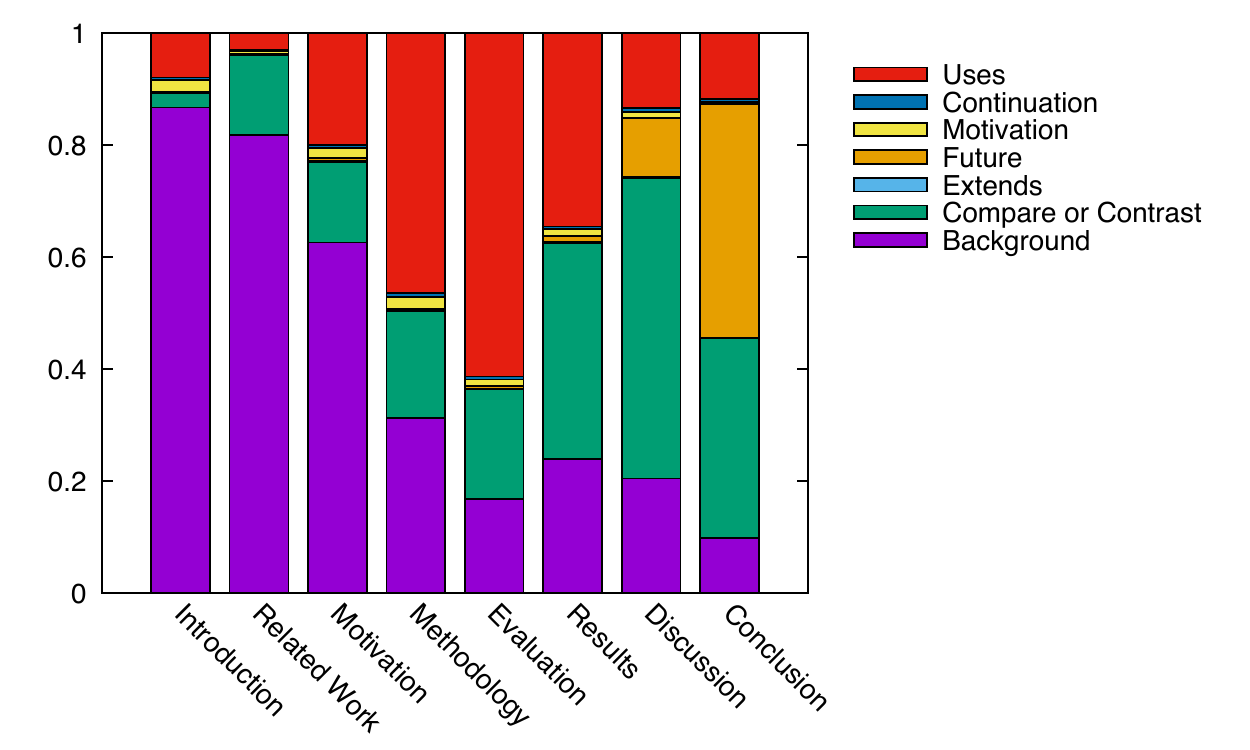}
    \caption{Expected percentage of citation functions per section shows a clear narrative trajectory across sections.  }
    \label{fig:funcs-per-sec}
\end{figure}

\section{Narrative Structure of Citation Function}

In the next four sections, we apply the classifier trained on our combined dataset
(3083 citation function instances, 1969 instances for citation centrality)
to the ACL Anthology to study what citation roles can tell us about
scientific uptake and direction.

As a qualitative demonstration, we first examine the narrative structure of citation function across section.
Scientific papers commonly follow a structured narrative according to section: Introduction, Methodology, Results, and Discussion \cite{skelton1994analysis,nwogu1997medical}. Each part in the narrative adopts argumentative moves designed to convince the reader of the work's claims \cite{swales1986citation,swales1990genre}.  We  expect that this narrative is mirrored in how authors use their citations in sections,
with the section serving to frame the meaning of a citation \cite{goffman1974frame,Gumperz1982discourse}.

To test this hypothesis, the function classifier was applied to all 21,474 papers of the latest 2016 release of the ACL Anthology.  This yielded a dataset of 134,127 citations between papers in the ARC.
The resulting distributions of citation function (Figure \ref{fig:funcs-per-sec}), show that authors' citation usage 
indeed parallels the expected rhetorical framing: (1) establishing framing via
\textsc{Background} citations in the Introduction, Motivation, and Related Work sections to 
(2) introducing methodology with \textsc{Uses} citations in the Methodology and Evaluation sections, 
(3) a large increase in \textsc{Comparison or Contrast} for related literature in the Results and Discussions, and finally 
(4) closing comparisons and pointers to future directions. 
These trends also mirror the thematic structure identified in full-paper textual analyses \cite{skelton1994analysis,nwogu1997medical}.
By showing that a section contains citations serving a variety of functions, our findings further
point to a new direction for citation placement studies 
\cite{hu2013citations,ding2013distribution,bertin2016invariant},
which have largely treated all citations within a section as equivalent.

%% file: 4-venue.tex
\section{Venues and Citation Patterns}
\label{sec:full-data}

\begin{figure}[ht]
    \centering
%
        \centering
        \includegraphics[width=0.50\textwidth]{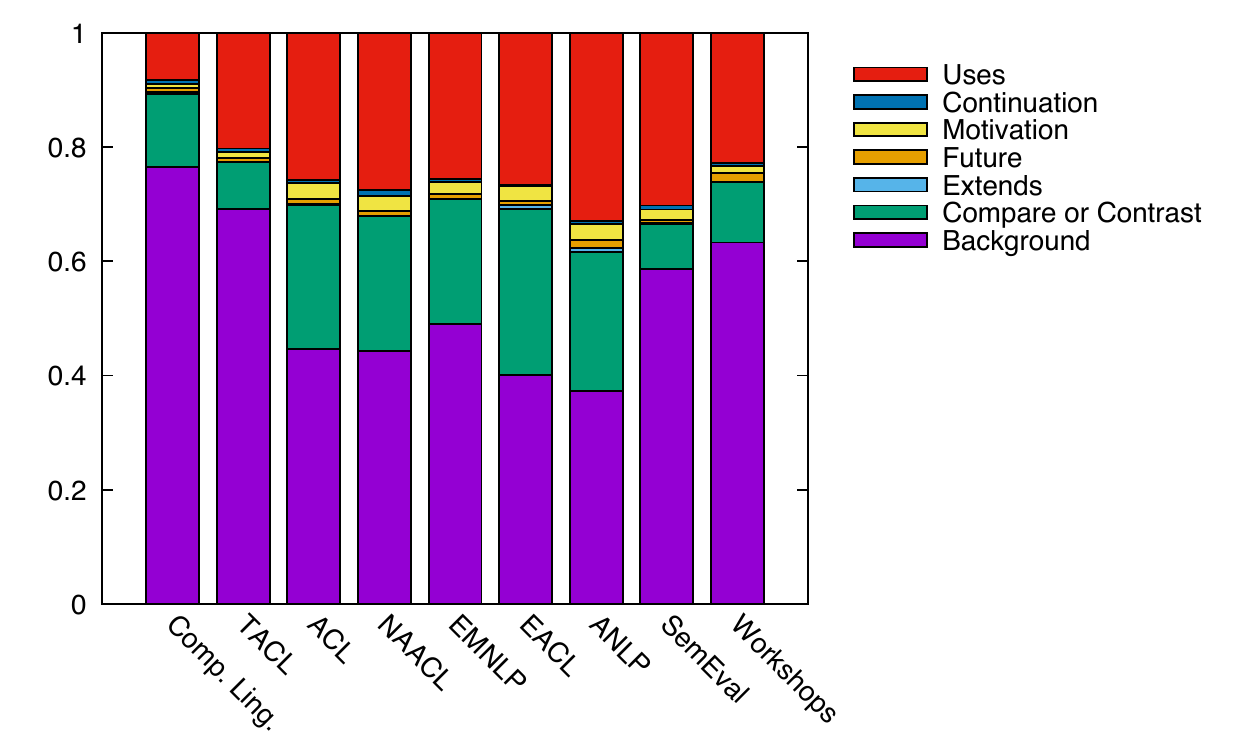}
   %
    \caption{Citation functions by venue.}
        \label{fig:types-per-venue}
\end{figure}

Does the venue in which a paper appears
affect the way it cites?  We used the same experimental setup as the previous section.
Figure \ref{fig:types-per-venue} shows 
citation function by venue for the 134,127 citations.

We find that similar venues have similar citation functions.
Journals have the highest percentage of \textsc{Background} citations,
suggesting that their extra space and wider temporal scope lends itself  more to positioning.
Conference venues  devote proportionally more space to contrast and comparison, presumably because
new NLP work is first presented at  conferences, hence requiring demonstrating the proposed technique is better than existing ones.
Workshops, by contrast, have relatively little comparison and instead 
use more \textsc{Background}; the experimental nature of workshop papers presumably results in fewer potential prospects for comparison.

%% file: 5-browsing.tex
\section{Citation Network Navigation}
    
The narrative in a scholarly work highlights salient aspects about
cited works for the reader.  As a result, a citation's framing may
motivate or discourage readers from seeking it
out. Understanding readers' citation-following behavior 
is an important part of understanding scholars' information gathering goals and strategies.
Yet while human navigation on hyperlinked networks such as Wikipedia have been explored extensively (e.g. \newcite{west2012human}),
we know very little about how scholars browse citation networks \cite{rouse1982human,bergstrom2006circleview}; and
no studies have looked at data from users' real-world behavior.
Do readers generally follow links to methodologies?  Are they more likely to follow links to essential papers?

The ACL Anthology offers an opportunity to examine these questions in detail,
since the server logs from the Anthology allow us to estimate when readers
follow citation links from one paper to another. We use this data to 
understand how the way a reference is cited is predictive of it being read. 

\noindent \textbf{Experimental Setup}~~
Paper requests were gathered from the complete Apache logs of the ACL Anthology website 
from November 2014 to September 2015.  After removing requests by bots and webspiders, the logs contain roughly 3.6M requests for 34,821 papers, with between 1 and 9588 requests per paper (mean 97.5).  
The access logs do not directly record unique user identifiers for tracking behavior over extended periods of time (e.g., as compared to session cookies); however, each request includes a string describing the user agent, which reports specific details on the operating system, browser, and version numbers. 
Because of the high number of unique combinations of user agent features and relatively low traffic to the website, these user agent strings are sufficient for identifying access patterns of individual users during short time intervals.
%
%
Activity traces are constructed by identifying sequential requests made by  the same user agent string for two or more papers, with no more than 60 minutes between requests.   The vast majority of traces consist of requests for two (80\%) or three papers (9.6\%).\footnote{To guard against cases where a user agent string might be used by multiple individuals in parallel, we remove all traces with 50 or more requests, though this is rare in practice (0.3\%).}


To compare with the observed human behavior, we construct a null model over the expected frequencies with which references of each type are visited.\footnote{
For each trace beginning with paper $p_i$ and continuing to one of its references $p_j \in R$ , the null model randomly selects a reference in $R$ with equal probability.   Visiting the same initial papers  seen in the observed data controls for differences in the number of references.  
The expected frequency distributions for reading behavior was created by generating visit counts from 500 simulations per trace with the null model and calculating the number of standard deviations away the observed frequency was from the expected frequency, i.e., the z-score.}
A positive z-score indicates that the reader accesses the paper more frequently than expected by chance. 

The effects of how a reference is cited is measured by calculating the z-score in visits with respect to references by class.  We treat a reference as \textsc{Essential} if any of its citations are labeled as such.   Because a reference may have multiple functions (e.g., both \textsc{Background} and \textsc{Uses}), 
we use a fractional count  proportional to the number of functions.

\begin{table}[t]
\small
\centering
    \sisetup{
    table-number-alignment = center,
    table-figures-integer = 2,
    table-figures-decimal = 2,
    table-space-text-pre = -,
    table-space-text-post = ***,
    }
  \begin{tabular}{
    S[table-text-alignment = left, table-column-width = 3.5cm]
    S
  }
        {\textbf{Citation Class}} & {\textbf{Z-score}} \\
        \hline
        {\textsc{Positional}} & -25.13*** \\
        {\textsc{Essential}}  & 25.13***  \\
        \hline
        {\textsc{Continuation}} &	41.73*** \\
        {\textsc{Extends}}       & 7.53*** \\
          {\textsc{Future}}     & 3.00*** \\
        {\textsc{Background}	}  & -0.17 \\
           {\textsc{Motivation} }   & -2.02* \\
        {\textsc{Uses}}	      & -3.10*** \\
        {\textsc{Compare orContrast}} & -4.25*** \\
    \end{tabular}
    \caption{The effects of reference centrality and function on whether a reader later accesses the paper.  Effects are measured using the Z-score between observed and expected access frequencies; * indicates significance at p $\le$ 0.05, *** at p $\le$ 0.01. }
    \label{tab:browsing}
    \vspace{-2mm}
\end{table}

\noindent \textbf{Results}~~
Centrality was strongly predictive of whether a reader would subsequently access the reference, shown in  Table \ref{tab:browsing} (top).  Confirming the hypothesis of \newcite{valenzuela2015identifying}, we find that \textsc{Essential} references are significantly more likely to be read than at chance (p $\le$ 0.01), while \textsc{Positioning}  are  less likely (p $\le$ 0.01). However, the effect of citation function (Table \ref{tab:browsing}, bottom) reveals a more complex story of reader behavior.  

Readers are  more likely to follow the temporal thread of an idea, reading the papers from which the current work originates (by others \textsc{Extends} or themselves \textsc{Continuation}) or how the current work could be applied in the future (\textsc{Future}).
However, readers are significantly less likely to follow \textsc{Uses} citations, despite these nearly always being \textsc{Essential}.
We speculate that when authors extend their work or the
work of others, they are more likely to omit details of the original
paper; hence, a reader must go back to the original paper to obtain a complete picture.
And because works within a research area often share 
methodologies and data (\textsc{Uses} citations),
a reader may be familiar with these references and not seek them out again.

By using the browsing patterns of an entire field, we have
demonstrated the strategies that scholars use to explore research.
Our findings more broadly motivate the need for scholarly search engines  like Semantic Scholar ({\small \tt{\href{https://www.semanticscholar.org/}{www.semanticscholar.org/}}}) to support browsing preferences by surfacing temporally-connected references instead of methodological citations.



%% file: 6-predict-impact.tex
\section{Predicting Future Impact}

The scholarly narrative told through citations provides the reader with support for its claims and technical competence \cite[p.\,34]{latour1987science}.
This framing could affect the perception of the work and, ultimately, how it is received and cited within the community
\cite{shi2010citing}.
Does the  narrative  a paper constructs through citation roles (the way it compares to related work,
or motivates, or points to the future) affect its reception?



\noindent \textbf{Experimental Setup}~~
To quantify the impact of citation usage, we test its ability to predict the cumulative number of citations a paper received within the first five years after publication, known to be highly representative of the eventual citation count \cite{wang2013quantifying,stern2014high}.
Our baseline model consists of
state of the art features for predicting citation count \cite{yan2011citation,yan2012better,chakraborty2014towards,dong2016can};
we then ask whether prediction is improved by adding the number of citations per function and centrality.

All papers with at least five years of publication history in the anthology were considered, yielding a set of 10,434 papers.
We used negative binomial models, which are more appropriate than  linear regression as citation counts are non-negative discrete counts,
and compared them using Akaike Information Criterion (AIC).

Five types of baseline features were included.
To model the amount of attention received by different research areas,
each paper is associated with its distribution over 100 topics, built using LDA over the ARC; to capture diversity, we include the entropy of the topic distribution.
We model the different publication venues via a categorical feature for the ACL venue in which the paper was published.
We include the publication year since the size of the field changes over time.
Multi-author papers are known to receive higher citation counts \cite{gazni2011investigating},
partially due to the effects of self-citation \cite{fowler2007does}, and therefore we include the number of authors on the paper.
Finally, we include the number of references.\footnote{To control for collinearity between citation-related predictors, we regress out the number of citations from the citation function counts and then in turn, regress out the citation function counts from citation centrality counts \cite{kutner2004applied,o2007caution}.  The resulting model has a variance inflation factor of $<$ 10 for all variables.}

\begin{table}
    \centering
    \small
    \begin{tabular}{lcc}
\textbf{Citation Class} & \textbf{Coefficient} & \textbf{p-value} \\\hline
{\textsc{Positioning}}    &        0.0423    &   0.124      \\
{\textsc{Essential}}      &       0.0644   &    0.025     \\
\hline
{\textsc{Motivation }}    &       0.0471   &    0.035 \\
{\textsc{Continuation }}  &        -0.0004~ &     0.994 \\
{\textsc{Future}}     &           0.0086   &    0.778 \\
{\textsc{Uses}}         &         0.0400  &     0.000 \\
{\textsc{Background}}   &         0.0195   &    0.045 \\
{\textsc{Extends}}     &         -0.0403~  &    0.557 \\ 
{\textsc{Compare or Contrast}} &  0.0253  &     0.010    \\

    \end{tabular}
  \caption{Regression coefficients for roles when predicting the number of citations within 5 years of publication.
  }
  \label{tab:regression}
\end{table}

\noindent \textbf{Results}~~
Knowledge of how a paper contextualizes itself indeed helps predict its future impact;  citation function and centrality  each improve AIC model fit 
(likelihood ratio test,  p $\le$ 0.01 for each), and adding them together improves AIC again (p $\le$ 0.01).

Four main insights can been seen through which types of citations are significantly predictive of higher impact (p$\le$0.05), shown in Table \ref{tab:regression}.
First, papers that closely integrate more works as essential to their contribution are cited more than those using more citations to position their work.  This trend parallels the effects seen with \textsc{Uses} citations, which suggests that papers  that leverage existing methodologies and data have higher impact.
Second, citations to motivating works have the most impact on future citation.  We speculate that such citations provide the reader with an explicit signal of external support for a line of research which inspires future work in the area as a result.
Third, \newcite[p.\,54]{latour1987science} has suggested that authors may deflect criticism of their work (improving its perception) by claiming it as an extension, rather than comparing it with prior work.  However, we find that additional comparison increases impact, while no significant effect was seen for \textsc{Extends}.  
We speculate that \textsc{Extends} and \textsc{Continuation} citations could lead to the work being perceived as incremental, thereby reducing its novelty and anticipated impact.
Fourth, citations indicating a temporal connection had no significant effect on citation frequency (p$>$0.05)---despite our earlier results showing that these citations are more likely to be read.

%% file: 7-rapid.tex
\section{The Growth of Rapid Discovery Science}

\begin{figure*}[t]
\hspace*{-10pt}
    \begin{subfigure}[t]{0.32\textwidth}
        \centering
        \includegraphics[height=4.7cm]{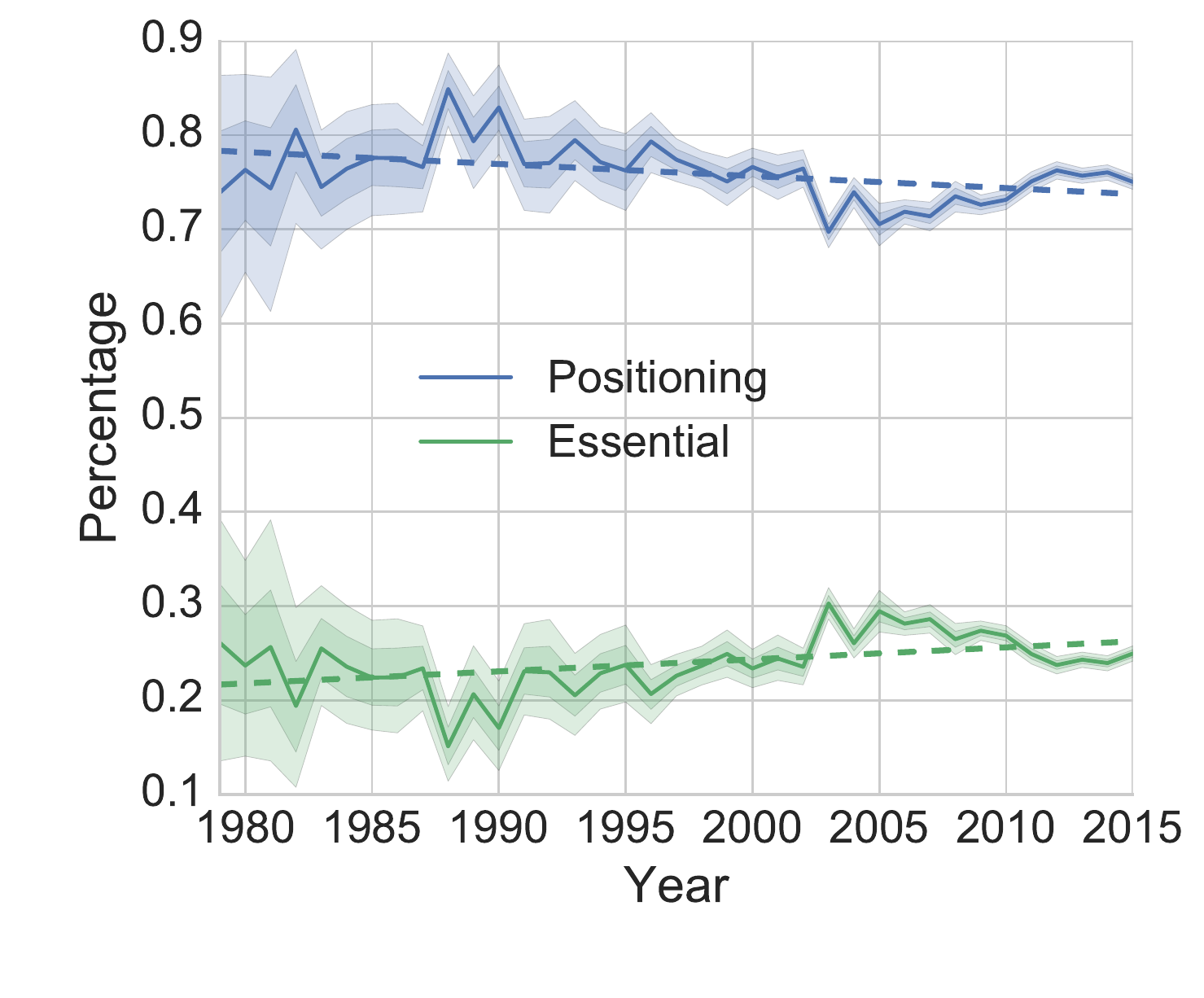}
        \caption{Centrality}
        \label{fig:type-change}
    \end{subfigure}%
    ~ 
    \begin{subfigure}[t]{0.32\textwidth}
        \centering
        \includegraphics[height=4.7cm]{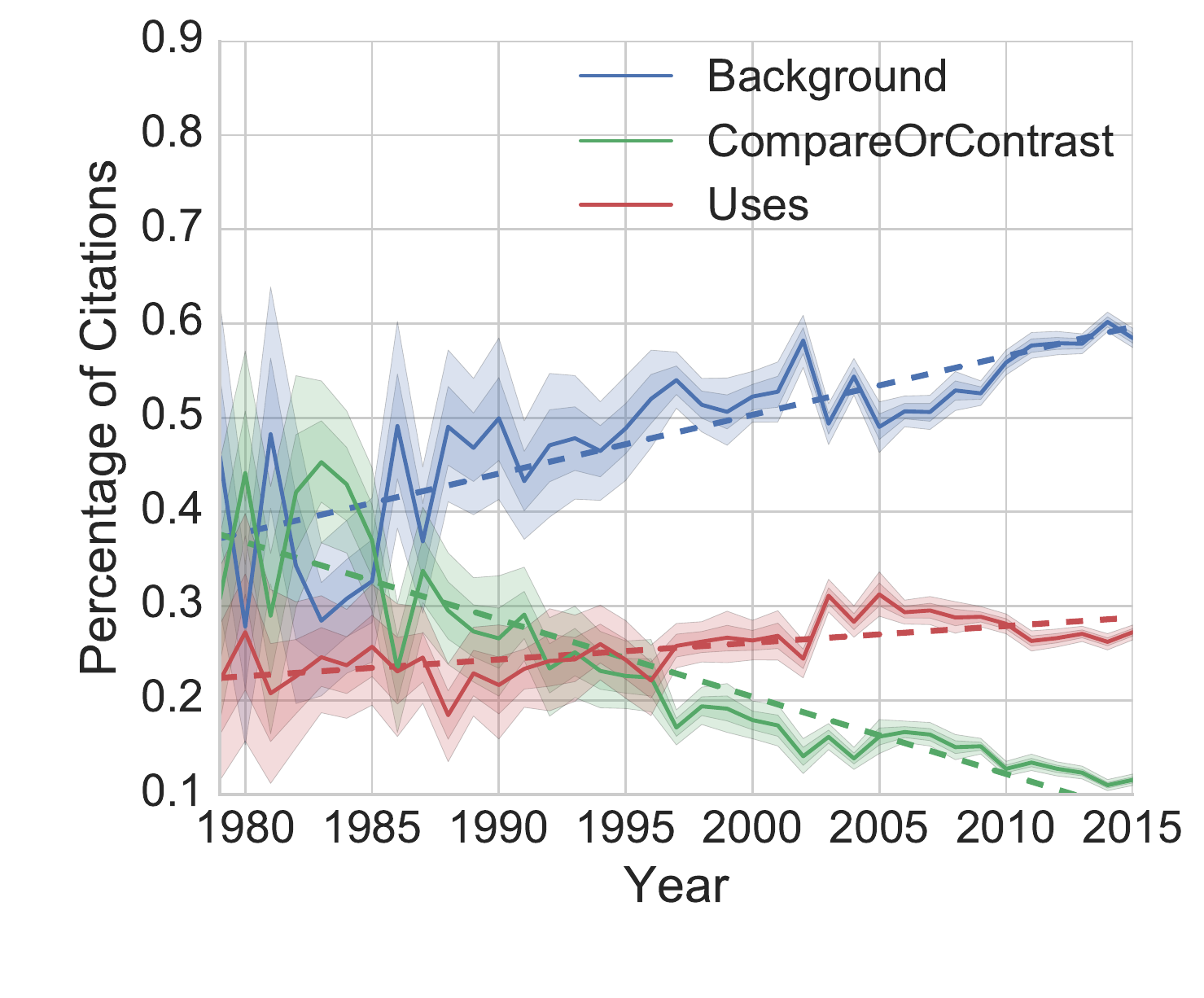}
        \caption{Function}
        \label{fig:func-change}
    \end{subfigure}  
     ~ 
    \begin{subfigure}[t]{0.32\textwidth}
        \centering
        \includegraphics[height=4.5cm]{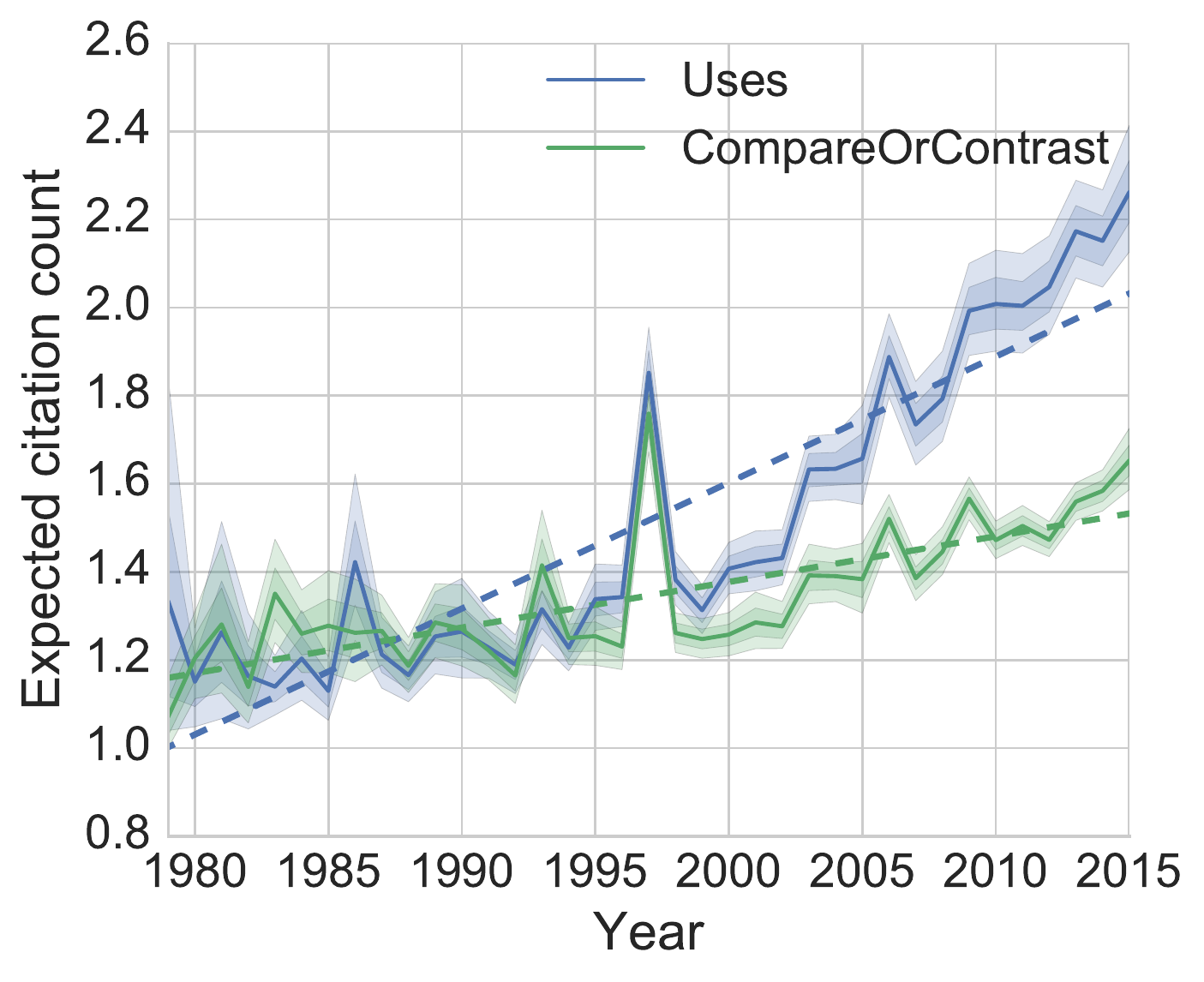}
        \caption{Expected number of citations}
        \label{fig:cite-change}
    \end{subfigure}    
    \caption{Changes in the percentage of citation types (\ref{fig:type-change}) and functions (\ref{fig:func-change}) per paper reveals an increase in the overall percentage of \textsc{Essential} references per paper and decrease in the percentage of \textsc{Comparison or Contrast}.  However, the average cited paper receives an increasing number of   \textsc{Uses} and \textsc{Comparison or Contrast} citations per year (\ref{fig:cite-change}) showing that field increasingly builds upon the same set of papers, providing a methodological lineage.  }
    \label{pe-change}
\end{figure*}

As scientific fields evolve, new subfields are often initially based around method technology, 
which starts a community discussion of the best way to use and improve the technology \cite{moody2004structure}.
NLP has witnessed the emergence of several such subfields from the early 
grammar based approaches in the 1950s-1970s to the statistical revolution in the 1990s to the recent deep learning models \cite{jones2001natural,anderson12}.  
\newcite{collins1994social} proposed that a field can undergo a particular shift, to
what he calls {\em rapid discovery science}, when the field (a)  reaches high consensus on research topics and methodologies, and (b) develops genealogies of method technologies that build upon one another.
As a result of increased consensus, a field emphasizes moving to new research problems rather than contesting the results of prior approaches.
Collins  claims this shift characterizes natural sciences, but not many social sciences, which are instead more likely to  emphasize
this process of contesting.

We propose to measure the two facets of rapid discovery science by quantifying the degrees to which authors (a)  defend their work through positioning and comparison, with lower rates suggesting more consensus, and (b)  focus on the same works for methodology and comparison, with increased rates signaling the development of methodological lineages.
We propose that the increased use of shared evaluations, and the statistical methodology borrowed originally from electrical engineering
\cite{hall08} has led NLP to undergo a shift towards rapid discovery science.

\noindent \textbf{Experimental Setup}~~
The expected frequencies of citation functions and centrality  were determined using the fully-classified ACL Anthology network.  Due to variance in the number of papers per year, we report bootstrapped confidence intervals in the expected percentage of each at 68\% and 95\%.

\noindent \textbf{Results}~~
The NLP field shows a significant increase in consensus consistent with the rise in rapid discovery science, evidenced through two main trends.  

First, NLP authors use a decreasing number of \textsc{Positioning} references  (Pearson's $r$= -0.446, p $\le$ 0.01)---Figure \ref{fig:type-change}---with an even sharper decrease in the number of works used for comparison and contrast ($r$= -0.899, p $\le$ 0.01)---Figure \ref{fig:func-change}.  Instead of comparing to others, it seems that authors simply acknowledge prior work as \textsc{Background}. Despite an increase in \textsc{Background} citations, the total percentage of comparative and background citations (the two main Positioning functions; see Table 2) still declines ($r$= -0.663, p $\le$ 0.01), with authors instead increasingly including more \textsc{Uses} citations.
\newcite[p. 50]{latour1987science} argues that \textsc{Positioning} references are critical to an author's defense of an idea.
We therefore interpret the observed decrease in \textsc{Positioning} references as signaling a reduced need for authors to defend aspects of their work.  Authors are able to compare against fewer papers  due to the fields growing consensus on the validity of the problem and methodological contribution.

Note that there is a small but significant increase in the number of \textsc{Positioning} references between  2009 and 2011.
This transition corresponds to the date at which ACL venues began allowing unlimited references (2010 for ACL, 2011 for NAACL, etc.).
Unlimited extra space for citations acted to distort the framing of citations; given unlimited space,
authors chose to include proportionally more \textsc{Positioning} citations.\footnote{Note that this change acts against the general
decrease in \textsc{Positioning}; considering only 1980-2009, the decrease in \textsc{Positioning} is even larger
($r$= -0.568, p $\le$ 0.01 ).}

In the second trend, authors are more likely to use and compare
against the same papers, as shown in Figure \ref{fig:cite-change}
by the rise in expected incoming citations to those works compared
against ($r$= 0.734, p $\le$ 0.01) and used ($r$=0.889, p $\le$
0.01).
For example, in 1991, authors compared with a diffuse group of parsing papers, e.g., \cite{shieber1988uniform,pereira1983parsing,haas1989parsing}, with such papers receiving at most three citations that year; whereas in 2000, most comparisons were to a core set of parsing papers, e.g., \cite{collins2003head,W99-0629,collins1997three}, with a much sharper (lower entropy) distribution of citations.  
These trends show the increased incorporation of prior work to form a lineage of method technologies as well as show increased consensus on which works are sufficient for comparing against in order to establish a claim.
%
%
These results also empirically confirm the  observation of \newcite{jones2001natural} that a major trend in NLP in the 1990s 
was an increase in reusable technologies and evaluations, like the BNC \cite{leech1992100} and the Penn Treebank \cite{marcus1993building}.

More broadly, our work points to the future of NLP as a quickly moving field of high consensus and suggests that artifacts that facilitate consensus such as shared tasks and open source research software will be necessary to continue this trend.

%% file: 9-conclusion.tex
\section{Conclusion}
\vspace{-0.5mm}

Not all citations are created equal.  Using a new corpus annotated with citation function and centrality, we develop a state-of-the-art classifier, demonstrating the importance of  novel unsupervised features related to topic models and argument structure, and label all the citations for an entire field (with all data and materials released at {\footnotesize \url{https://github.com/}\emph{anon}}).

We then show that citation function and centrality reveal salient behaviors of writers, readers, and the field as a whole:
(1) authors are sensitive to discourse structure and venue when citing,
(2) readers follow temporal threads of ideas in their browsing of citation networks, rather than methodological threads,
(3) the way in which an author relates their work to the literature is predictive of the number of citations its receives, with the community favoring well-motivated works that integrate and compare with other works, 
and (4) the NLP field as a whole has seen increased consensus in what constitutes valid work---with a reduced need for positioning and excessive comparison---demonstrating its shift towards rapid discovery science.